\documentclass[journal]{IEEEtran}
\newcommand{\etal}{\textit{et al}. }
\newcommand{\ie}{\textit{i}.\textit{e}. }
\newcommand{\eg}{\textit{e}.\textit{g}. }

\usepackage{times}
\usepackage{epsfig}
\usepackage{graphicx}
\usepackage{amssymb}
\usepackage{multirow}
\newcommand{\argmin}{\operatornamewithlimits{arg\ min}} % thin space, limits on side in displays
\usepackage[linesnumbered,ruled]{algorithm2e}
\usepackage{ctable} % for \specialrule command
\usepackage{lineno}
\usepackage[pagebackref=true,breaklinks=true,letterpaper=true,colorlinks,bookmarks=false]{hyperref}

\ifCLASSINFOpdf
\else
\fi
\usepackage[cmex10]{amsmath}
\begin{document}
\title{Knowledge Projection for Effective Design of Thinner and Faster Deep Neural Networks}

\author{Zhi~Zhang,
        Guanghan~Ning,
        and~Zhihai~He% <-this % stops a space
\thanks{Z. Zhang, G. Ning and Z. He are with the Department
of Electrical and Computer Engineering, University of Missouri, Columbia,
MO, 65203 USA.}}% <-this % stops a space
\maketitle

\begin{abstract}
While deeper and wider neural networks are actively pushing the performance limits of various computer vision and machine learning tasks, they often require large sets of labeled data for effective training and suffer from extremely high computational complexity. 
In this paper, we will develop a new framework for training deep neural networks
on datasets with limited labeled samples using cross-network knowledge projection which is able to improve the network performance while reducing the overall computational complexity significantly. 
Specifically, a large pre-trained teacher network is used to observe samples from the training data. A projection matrix is learned to project 
this teacher-level knowledge and its visual representations from an intermediate layer of the teacher network to an intermediate layer of a thinner and faster student network to guide and regulate its training process. 
Both the intermediate layers from the teacher network and the injection layers
from the student network are adaptively selected during training by evaluating a joint loss function in an iterative manner. 
This knowledge projection framework allows us to use crucial knowledge learned by large networks to guide the training of thinner student networks, 
avoiding over-fitting, achieving better network performance, and significantly reducing the complexity.
Extensive experimental results on benchmark datasets have demonstrated that our proposed knowledge projection approach outperforms existing methods, improving accuracy by up to 4\% while reducing network complexity by 4 to 10 times, which is very attractive for practical applications of deep neural networks.
\end{abstract}


\renewcommand{\baselinestretch}{2.1}

% \fi
\IEEEpeerreviewmaketitle

\section{Introduction}

\IEEEPARstart{R}{ecently}, large neural networks have demonstrated extraordinary performance on various computer vision and machine learning tasks. Visual competitions on large datasets such as ImageNet \cite{ILSVRC15} and MS COCO \cite{MSCOCO} suggest that \textit{wide} and \textit{deep} convolutional neural networks tend to achieve better performance, if properly trained on sufficient labeled data with well-tuned hyper-parameters, at the cost of extremely high computational complexity. Over-parameterization in large networks seems to be beneficial for the performance improvement \cite{denil2013predicting, hinton2012improving}, however,
the requirements for large sets of labeled data for training and high computational complexity pose significant challenges for us to develop and deploy deep neural networks in practice. 

First, low power devices such as mobile phones, cloud based services with high throughput demand, and real-time systems, 
have limited computational resources, which requires that the network inference or testing should have low computational complexity. 
Besides the complexity issue, a large network often consumes massive storage and memory bandwidth. Therefore, smaller and faster networks are often highly desired  in real-world applications.
Recently, great efforts have been made to address the network speed issue. A variety of model compression approaches \cite{NIPS1989_250,hassibi1993second,han2015deep,jaderberg2014speeding,hinton2015distilling} were proposed to obtain faster networks that mimic the behavior of large networks. 
Second, in practical applications,  we often have access to very limited labeled samples. It is very expensive to obtain human labeled ground-truth samples for training. In some applications domains, 
it is simply not feasible to accumulate enough training examples for deep networks  \cite{pan2011domain,zhang2013domain,wang2014flexible,tzeng2015simultaneous} 
.

Interestingly, these two problems are actually coupled together. The network capacity is often positively correlated to its task complexity. For instance, we would expect a small network classifier of two classes (\eg, dog and cat) to achieve a similar  level of accuracy as a significantly larger network for tens of thousand classes of objects. Existing solutions to obtaining a fast network on new tasks is often based on a two-step approach: train the network on a large dataset, then apply model compression or distillation to the network after fine-tuning or transfer learning on the new dataset \cite{hinton2015distilling}. 
Each step is performed separately and they are not jointly optimized.
Therefore, how to jointly address the problems of network compression, speed up, and domain adaptation becomes a very important and intriguing research problem. 

\begin{figure}[h]
\begin{center}
   \includegraphics[width=0.9\linewidth]{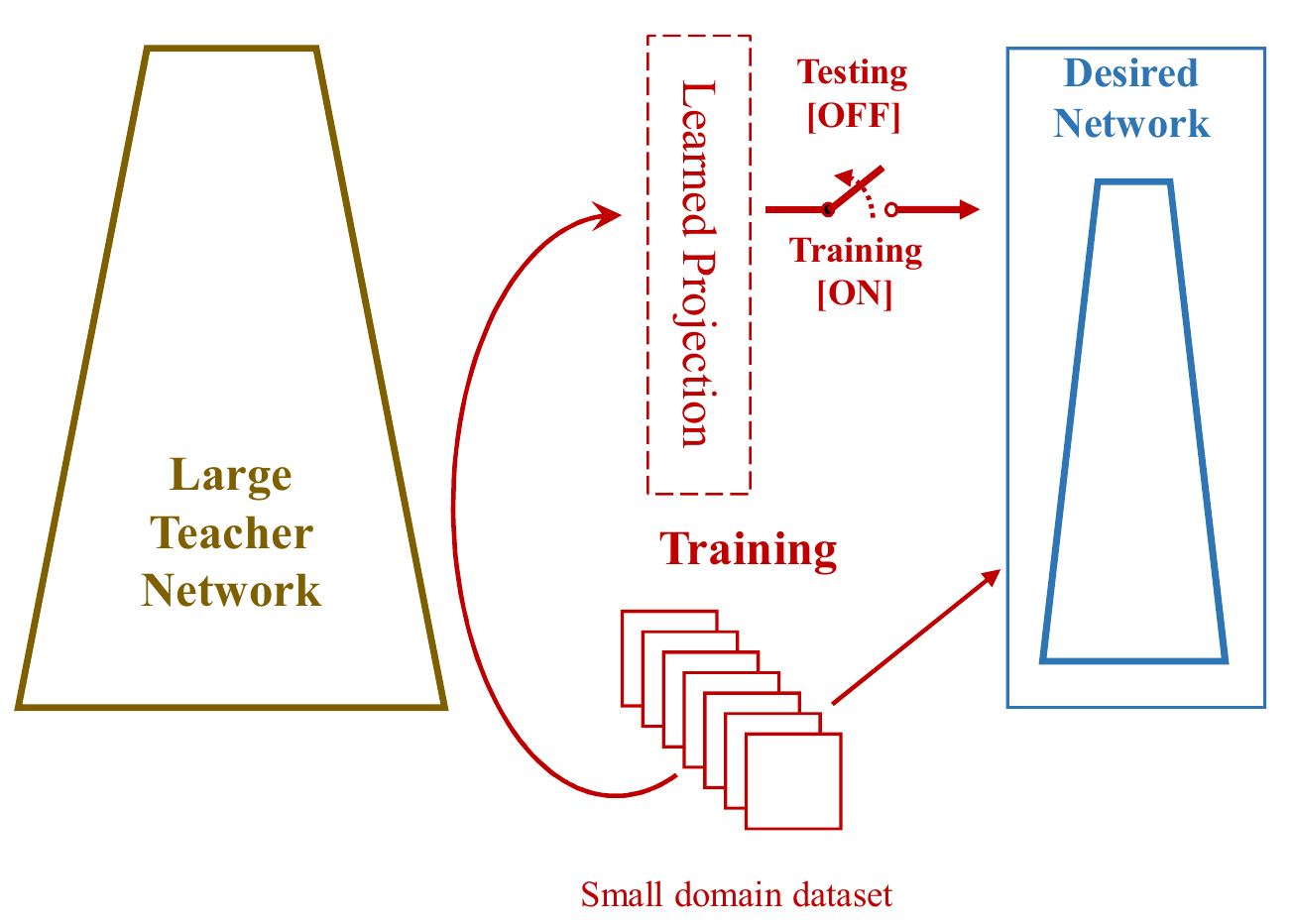}
\end{center}
\caption{System overview. We apply learned projection during training to guide a standard thinner and faster network for inference on a smaller domain dataset.}
\label{fig:overview}
\end{figure}

A successful line of work \cite{bucilua2006model,hinton2015distilling,romero2014fitnets, hou2016skeleton, xiong2016convolutional} suggest that cumbersome large neural networks, despite their redundancy, have very robust interpretation of training data. By switching learning targets from labels to interpreted features in small networks, we have observed not only speed-ups but also performance improvements. Inspired by this phenomenon, we are interested to explore if this interpretation power is still valid across different (at least similar) domains, and to what extent of performance a newly trained student network can achieve with the help of a large model pre-trained on different datasets. 

In this paper, we propose a Knowledge Projection Network (KPN) with a two-stage joint optimization method for training small networks under the guidance of a pre-trained large teacher network, as illustrated in Figure \ref{fig:overview}. In KPN, a knowledge projection matrix is learned to extract distinctive representations from the teacher network, and used to regularize the training process of the student network. We carefully design the teacher-student architecture and joint loss function so that the smaller student network can benefit from extra guidance while learning towards specific tasks. Our major observation is that, by learning  necessary representations from a teacher network which is fully trained on a large dataset, a student network can disentangle the explanatory factors of variations in the new data and achieve more precise representation of the new data from a smaller number of examples. Thus, same level performance can be achieved using a smaller network. 
Extensive experimental results on benchmark datasets have demonstrated that our proposed knowledge projection approach outperforms existing methods, improving accuracy by up to 4\% while reducing network complexity by 4 to 10 times, which is very attractive for practical applications of deep neural networks.

Our contributions in this paper are summarized as follows:
(1) we propose a new architecture to transfer the knowledge from a large  teacher network pre-trained on a large dataset into a thinner and faster student network to guide and facilitate its training on a smaller dataset. Our approach addresses the issues of network adaptation and model compression at the same time. 
(2) We have developed a method to learn a projection matrix which is able to project the visual features from the teacher network into the student network to guide its training process and improve its overall performance.  
(3) We have developed an iterative method to select the optimal path for knowledge projection between the teacher and student networks. 
(4) We have implemented the proposed method in MXNet  and conducted extensive experiments on benchmark datasets to demonstrate that our method is able to significantly reduce the network computational complexity by 4-10 times while largely maintaining or even improving the network performance by a significant margin. 

The rest of this paper is organized as follows. Related work is reviewed in Section \ref{sec:related}. We present the proposed Knowledge Projection Network in Section \ref{sec:kpn}. Experimental results are presented in Section \ref{sec:exp}. Finally, Section \ref{sec:con} concludes this paper.

\section{Related Work}
\label{sec:related}
Large neural networks have demonstrated extraordinary performance on various computer vision and machine learning tasks. During the past a few years, researchers have been 
investigating how to deploy these deep neural networks in practice.
There are two major problems that need to be carefully addressed: the high computational complexity of the deep neural network and the large number labeled samples required to train the network \cite{kim2009configurable, sudha2011self}.  
Our work is closely related to domain adaptation and model compression, which are reviewed in this section.

To address  the problem of inadequate labeled samples for training, methods for network domain adaptation \cite{pan2010survey, wang2014flexible, long2015learning}
have been developed, which enable learning on new domains with few labeled samples or even unlabeled data. Transfer learning methods have been proposed over the past several years, and we focus on \textit{supervised learning} where a small amount of labeled data is available. It has been  widely recognized that the difference in the distributions of different domains should be carefully measured and reduced \cite{long2015learning}. Learning shallow representation models to reduce domain discrepancy is a promising approach, however, without deeply embedding the adaptation in the feature space, the transferability of shallow features will be limited by the task-specific variability. Recent transfer learning method coupled with deep networks can learn more transferable representations by embedding domain adaptations in the architecture of deep learning \cite{ganin2014unsupervised} and outperforms traditional methods by a large margin. Tzeng \etal \cite{tzeng2015simultaneous} optimizes domain invariance by correcting the marginal distributions during domain adaptation. The performance has been improved, but only within a single layer. Within the context of deep feed-forward neural networks, \textit{fine-tune} is an effective and overwhelmingly popular method \cite{zeiler2014visualizing, oquab2014learning}. Feature transferability of deep neural networks has been comprehensively studied in \cite{yosinski2014transferable}. It should be noted  that this method does not apply directly to many real problems due to insufficient labeled samples in the target domain. There are also some shallow architectures \cite{ajakan2014domain, ghifary2014domain} in the context of learning domain-invariant features. Limited by representation capacity of shallow architectures, the performance of shallow networks is often inferior to that 
of deep networks \cite{long2015learning}. 

With the dramatically increased demand of computational resources by deep neural networks, there have been considerable efforts to design smaller and thinner networks from larger pre-trained network in the literature. 
A typical approach is to prune unnecessary parameters in trained networks while retaining similar outputs. Instead of removing close-to-zero weights in the network, LeCunn \etal proposed Optimal Brain Damage (OBD) \cite{NIPS1989_250} which uses the second order derivatives to find trade-off between performance and model complexity. Hassibi \etal followed this work and proposed Optimal Brain Surgeon (OBS) \cite{hassibi1993second} which outperforms the original OBD method, but was more computationally intensive. Han \etal \cite{han2015learning} developed a method to prune state-of-art CNN models without loss of accuracy. Based on this work, the method of deep compression \cite{han2015deep} achieved better network compression ratio using ensembles of parameter pruning, trained quantization and Huffman coding, achieved 3 to 4 times layer-wise speed up and reduced the model size of VGG-16 \cite{simonyan2014very} by 49 times. This line of work focuses on pruning unnecessary connections and weights in trained models and optimizing for better computation and storage efficiency.

Various factorization methods have also been proposed to speed up the computation-intensive matrix operations which are the major computation in the convolution layers. For example,  methods have been developed to use matrix approximation to reduce the redundancy of weights. Jenderberg \etal \cite{jaderberg2014speeding} and Denton \etal \cite{denton2014exploiting} use SVD-based low rank approximation. For example, Gong \etal \cite{gong2014compressing} use a clustering-based product quantization to reduce the size of matrices by building an indexing. Zhang \etal \cite{zhang2016accelerating} successfully compressed very deep VGG-16 \cite{simonyan2014very} to achieve 4 times speed up with 0.3\% loss of accuracy based on Generalized Singular Value Decomposition and special treatment on non-linear layers. This line of approaches can be configured as data independent processes, but fine-tuned with training data to improve the performance significantly. In contrast to off-line optimization, Ciresan \etal \cite{cirecsan2011high}	trained a sparse network with random connections, providing good performance with better computational efficiency than densely connected networks.

Rather than pruning or modifying parameters from existing networks, there has been another line of work in which a smaller network is trained from scratch to mimic the behavior of a much larger network. Starting from the work of Bucila \etal \cite{bucilua2006model} and  Knowledge Distillation (KD) by Hinton \etal \cite{hinton2015distilling}, the design of smaller yet efficient networks has gained a lot of research interest. Smaller networks can be shallower (but much wider) than the original network,  performing as well as deep models, as shown by Ba and Caruna in \cite{ba2014deep}. The key idea of knowledge distillation is to utilize the internal discriminative feature that is implicitly encoded in a way not only beneficial to original training objectives on source training dataset, but also has a side-effect of eliminating incorrect mappings in networks. It has been demonstrated in \cite{hinton2015distilling} that small networks can be trained to generalize in the same way as large networks with proper guidance. FitNets \cite{romero2014fitnets} achieved better compression rate than knowledge distillation by designing a deeper but much thinner network using trained models. The proposed hint-based training is one step further beyond knowledge distillation  which uses a finer network structure. Nevertheless, training deep networks has proven to be challenging \cite{erhan2009difficulty}. Significant efforts have been devoted to alleviate this problem. Recently, adding supervision to intermediate layers of deep networks is explored to assist the training process \cite{lee2015deeply, szegedy2015going}. These methods assume that source and target domains are consistent. It is still unclear whether the guided training is effective when 
the source and target domains are significantly different.

In this paper, we consider a unique setting of the problem. We use a  large network pre-trained on a large dataset (\eg, the ImageNet) to guide the training of a thinner and faster network on a new smaller dataset with limited labeled samples, involving  adaptation over different data domains and model compression at the same time. 

\section{Knowledge Projection Network}
\label{sec:kpn}

In this section, we present the proposed Knowledge Projection Network (KPN). We start with the KPN architecture and then explain the knowledge projection layer design. A multi-path multi-stage training scheme coupled with iterative pruning for projection route selection is developed afterwards.

\subsection{Overview}

\begin{figure*}[ht]
\begin{center}
   \includegraphics[width=0.99\linewidth]{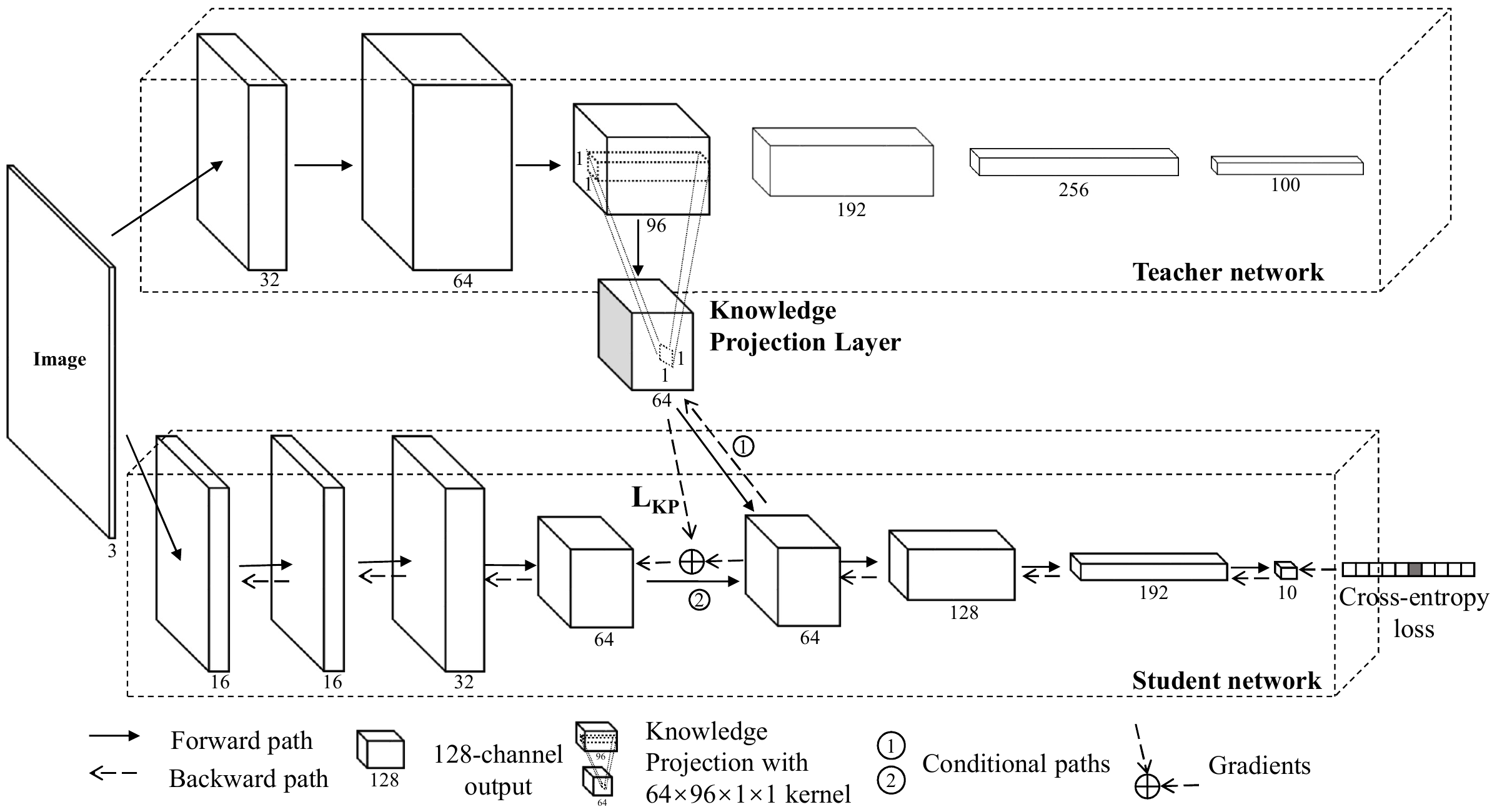}
\end{center}
   \caption{KPN architecture. Solid arrows showing the forward data-flow, dotted arrows showing the paths for gradients.}
\label{fig:architecture}
\end{figure*}

An example pipeline of KPN is illustrated in Figure \ref{fig:architecture}. Starting from a large teacher network pre-trained on a large dataset, a student network is designed to predict desired outputs for the  target problem with guidance from the teacher network. The student network uses similar buiding blocks as the teacher network, such as Residue \cite{he2016deep}, Inception \cite{szegedy2016rethinking} or stacks of plain layers \cite{simonyan2014very},  sub-sampling and BatchNorm \cite{ioffe2015batch} layers. The similarity in baseline structure ensures smooth transferability. Note that the convolution layers consume most of the computational resources.
Their complexity can be modeled by the following equation

\begin{equation}
\label{eq:multi_add}
    {\cal C} = \sum_{i=1}^{N-1}{C_i \cdot H_i \cdot W_i \cdot C_{i+1} \cdot K^H_{i} \cdot K^W_{i}},
\end{equation}

where the computational cost is multiplicatively related to the number of input $C_i$ and output channels $C_{i+1}$, the spatial size of input feature map $H_i \cdot W_i$ where $H_i$ and $W_i$ are the height and width of the feature map at the $i$-th layer, and kernel size $K^H_{i} \cdot K^W_{i}$. 
The student network is designed to be thinner (in terms of filter channels) but deeper to effectively reduce network capacity while preserves enough representation power \cite{ba2014deep, romero2014fitnets}. We depict the convolutional blocks in Figure \ref{fig:bottleneck} that are used to build the \textit{thin} student networks. In contrast to standard convolutional layers, a squeeze-then-expand \cite{iandola2016squeezenet, redmon2016yolo9000} structure is effective in reducing the channel-wise redundancy by inserting spatially narrow ($1\times1$) convolutional layers between $3\times3$ standard convolutional layers. We denote this structure as bottleneck Type A and extend it to a more compact squeeze-expand-squeeze shape, namely bottleneck Type B. With (\ref{eq:multi_add}), we can calculate the proportional layer-wise computation cost for the standard convolutional layer, bottleneck Type A and B, respectively. For simplicity, feature map dimensions are denoted in capital letters, and we use identical size for kernel height and width, denoted as $K$, without loss of generality:

\begin{figure}[h]
\begin{center}
\includegraphics[width=0.85\linewidth]{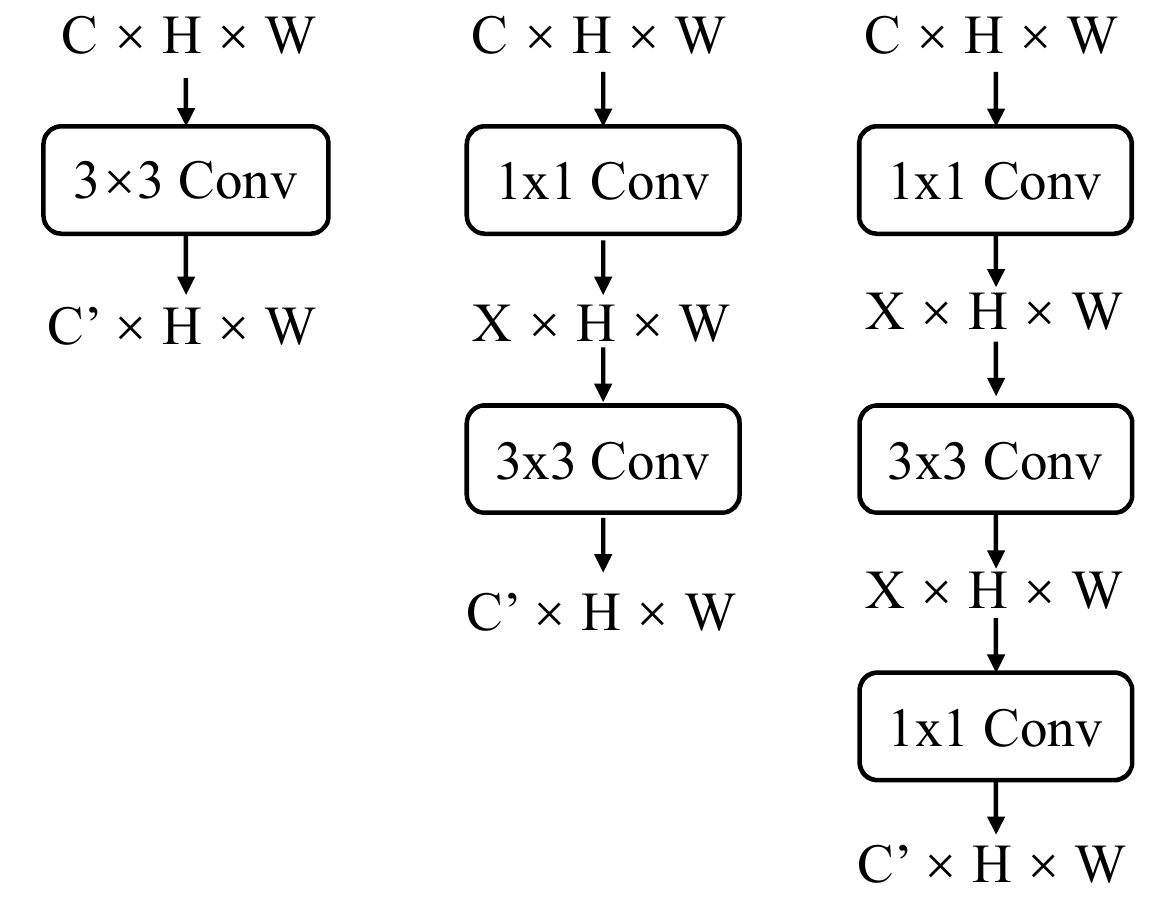}
\end{center}
\caption{Left: Standard 3x3 Convolutional layer. Middle: Bottleneck type A. Right: Bottleneck type B. $H$ and $W$ are feature spatial height and width, $C, X, C'$ are input, reduced and output channels for this building block, respectively. For simplicity, batch-norm and activation layers are omitted in this figure.}
\label{fig:bottleneck}
\end{figure}

\begin{equation}
\label{eq:multi_add_starndard}
    {\cal C}_{standard} = C \cdot H \cdot W \cdot C' \cdot K^2,
\end{equation}

\begin{equation}
\label{eq:multi_add_a}
    {\cal C}_{TypeA} = C \cdot H \cdot W \cdot X + X \cdot H \cdot W \cdot C' \cdot K^2,
\end{equation}

\begin{equation}
\label{eq:multi_add_b}
    {\cal C}_{TypeB} = C \cdot H \cdot W \cdot X + X^2 \cdot H \cdot W \cdot K^2 + X \cdot H \cdot W \cdot C'.
\end{equation}

Combining (\ref{eq:multi_add_starndard}), (\ref{eq:multi_add_a}) and (\ref{eq:multi_add_b}),
we define the reductions in computation for Type A and B as

\begin{equation}
\label{eq:multi_add_reduction_a}
    \frac{{\cal C}_{TypeA}}{{\cal C}_{standard}} = \frac{X}{C' \cdot K^2} + \frac{X}{C} \approx \frac{X}{C},
\end{equation}

\begin{equation}
\label{eq:multi_add_reduction_b}
    \frac{{\cal C}_{TypeB}}{{\cal C}_{standard}} = \frac{X}{C' \cdot K^2} + \frac{X^2}{C\cdot C'} + \frac{X}{C \cdot K^2} \approx \frac{X^2}{C \cdot C'},
\end{equation}

Bottleneck structures A and B can effectively reduce the computational cost while preserve the dimension of feature map and receptive field, and the layer-wise reduction is controlled by $X$. For example, by cutting the bottleneck channels by half, \ie, $X = \frac{C}{2}$, we have the approximate reduction rate $ \frac{1}{2}$ for Type A, $\frac{1}{4} \sim \frac{1}{8}$ for Type B. In practice, the output channel $C'$ is equal to or larger than input channel $C$: $C'\in [C, 2C]$. We replace standard convolutional layers by bottleneck structures A and B in the teacher network according to computational budget and constitute corresponding student network. Layer-wise width multipliers $\alpha=\frac{X}{C}$ are the major contributor to model reduction. We use smaller $\alpha$ in deep layers where the feature is sparse and computational expensive layers where the gain is significant. The flexibility of bottleneck structures and elastic value range of $\alpha$ ensured we have enough degrees of freedom controlling the student network capacity.
In our KPN, the  student network is  trained by optimizing the following joint loss function:

\begin{equation}
\label{eq:loss_function}
    W^*_s = \argmin_{W_s} \lambda \cdot \mathcal{L}_{KP}(W_s, W_k) + \mathcal{L}_p(W_s) + \mathcal{R},
\end{equation}
where $\mathcal{L}_{KP}$ and $\mathcal{L}_p$ are loss from the knowledge projection layer and problem specific loss, respectively.  For example, for the problem-specific loss, we can choose the cross-entropy loss in many object recognition tasks. $\lambda$ is the weight parameter decaying during training, $W_k$ is the trained teacher network, $\mathcal{R}$ is a $L-2$ regularization term, and $W^*_s$ is the trained parameters in the student network. Unlike traditional supervised training, the knowledge projection loss $\mathcal{L}_{KP}$ plays an important role in guiding the training direction of KPN, which will be discussed in more detail in the following section.

\subsection{Knowledge Projection Layer Design}
In this work, the pre-trained teacher network and the student network analyze the input image simultaneously.
To use the teacher network to guide the student network, we propose to map the feature $\cal{F}_T$ of size $N$ learned at one specific layer of the teacher network into a feature vector  $\cal{F}_S$ of size $M$ and inject it into the student network to guide its training process.
For the mapping, we choose linear projection
\begin{equation}
\cal{F}_S = \cal{P} \cdot \cal{F}_T,
\end{equation} 
where $\cal{P}$ is an $N\times M$ matrix.
In deep convolutional neural networks, this linear projection matrix $\cal{P}$ can be learned by constructing a convolution layer between the teacher and student network. Specifically, 
we use a convolutional layer to bridge teacher's \textit{knowledge} layer and student's \textit{injection} layer. A \textit{knowledge} layer is defined as the output of a teacher's hidden convolutional layer responsible for guiding the student's learning process by regularizing the output of student's \textit{injection} convolutional layer. Let $O_h^t$, $O_w^t$ and $O_c^t$ be the spatial height, spatial width, and number of channels of the  knowledge layer output in the teacher network, respectively. Let $O_h^s$, $O_w^s$ and $O_c^s$ be the corresponding sizes of student's injection layer output, respectively. Note that there are a number of additional layers in the student network to further analyze the feature information acquired in the inject layer and contribute to the final network output.
We define the following loss function:
\begin{equation}
\label{eq:LKR}
    \mathcal{L}_{KP(W_s, W_k)} = h[\mu(x; W_k)] \cdot \left|r[\mu(x; W_k); W_{KP}] - v[x; W_s]\right|,
\end{equation}

\begin{equation}\label{ch:five:sec:5:eq4:1}
h(x) =
\begin{cases} 
      1, & \text{if x $\ge$ 0}, \\
      \eta, & \text{otherwise}
   \end{cases}
\end{equation}
where $\mu$ and $v$ represent the deep nested functions (stacks of convolutional operations) up to the knowledge and injection layer with network parameters $W_k$ and $W_s$, respectively. $r[\cdot ]$ is the knowledge projection function applied on $\mu[\cdot ]$ with parameter $W_{KP}$ which is another convolution layer in this work. $\mu$, $v$ and $r$ must be comparable in terms of spatial dimensionality.

The knowledge projection layer is designed as a convolutional operation with a $1\times1$ kernel in the spatial domain. As a result, $W_{KP}$ is a $O^t_c \times O^s_c \times 1 \times 1$ tensor. As a comparison, a fully connected adaptation layer will require $O^t_h \times O^t_w \times O^t_c \times O^s_h \times O^s_w \times O^s_c$ parameters which is not feasible in practice especially when the spatial size of output is relatively large in the early layers. Using the convolutional adaptation layer is not only beneficial for lower computational complexity, but also provides a more natural way to filter distinctive channel-wise features from the knowledge layers while preserve spatial consistency. The output of the knowledge projection layer will guide the training of student network by generating a strong and explicit gradient applied to backward path to the \textit{injection} layer in the following form

\begin{equation}
\Delta {W_{s,i}} = - \lambda \cdot \frac{\partial \mathcal{L}_{KP}}{\partial W_{s,i}},
\end{equation}
where $W_{s,i}$ is the weight matrix of injection layer in student network. 
Note that in (\ref{eq:LKR}), $h[\mu(x; W_k)]$ is applied to $\mathcal{L}_{KP}$ with respect to the hidden output of knowledge projection layer as a relaxation term. For negative responses from $\mu(x; W_k)$, $\mathcal{L}_{KP}$ is effectively reduced by the slope factor $\eta$, which is set to $0.25$ by cross-validation. Overall, $\mathcal{L}_{KP}$ acts as a relaxed $L1$ loss. Compared to $L2$ loss, $\mathcal{L}_{KP}$ is more robust to outliers, but still has access to finer level representations in $r[\mu(x; W_k); W_{KP}]$.

\begin{figure}[ht]
\begin{center}
   \includegraphics[width=0.9\linewidth]{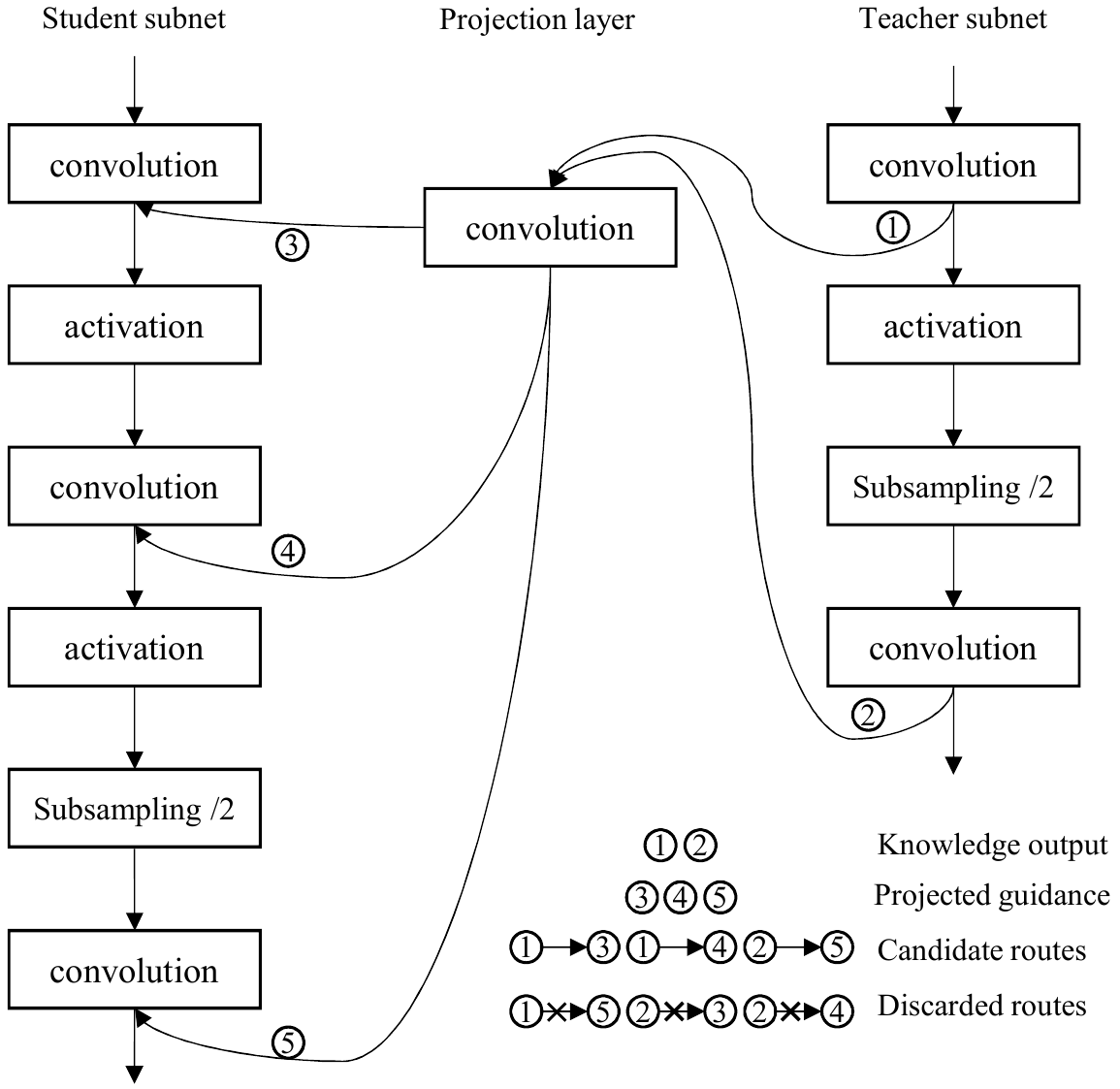}
\end{center}
   \caption{Candidate Routes of Knowledge Projection. Candidate routes are paths from teacher's knowledge layer to student's injection layer. Only one route will survive after iterative pruning.}
\label{fig:route_selection}
\end{figure}

%\begin{figure}[ht]
%\begin{center}
%   \includegraphics[width=0.99\linewidth]{figures/function.pdf}
%\end{center}
%   \caption{Choice of Knowledge Refining Route.}
%\label{fig:relax_function}
%\end{figure}

%%%%%%%%%%%%%%%%%%%%%%%%%%%%%%%%%%%%%%%%%%%%%%%%%

\subsection{Multi-Path Multi-Stage Training}
In the student network, layers after the \textit{injection} layer are responsible for adapting the projected feature to the final network output. This adaptation must be memorized throughout the training process. Those network layers before the injection layer aim to learn distinctive low-level features. Therefore, in our KPN framework, the student network and knowledge projection layer are randomized and trained in two stages: \textbf{initialization stage} and end to end \textbf{joint training stage}. 

%\subsubsection{initialization Stage}
In the initialization stage, Path \textcircled{2} in Figure \ref{fig:architecture} is disconnected, \ie the knowledge projection layer together with the lower part of student network is trained to adapt the intermediate output of teacher's knowledge layer to the final target by minimizing $\mathcal{L}_p$, which is the loss for target task, e.g., softmax or linear regression loss.  The upper part of student network is trained sorely by minimizing $\mathcal{L}_{KP}$. In this stage, we use the projection matrix as an implicit connection between upper and lower parts in the student network. The upper student network layers are always optimized towards features interpreted by the projection matrix, and have no direct access to targets. This strategy prevents the student network from over-fitting quickly during the early training stage which is very hard to correct afterwards.

%\subsubsection{End to End Joint Training Stage}
After the initialization stage, we then disconnect Path \textcircled{1} and reconnect Path \textcircled{2}, the training now involves jointly minimizing the objective function described in (\ref{eq:loss_function}). 
Using the results from stage 1 as the initialization, the joint optimization process aims to establish smooth transitions inside the student network from the input to the final output. The loss $\mathcal{L}_{KP}$ injected into the student network continues to regularize the training process. In this way, the student network is trained based on a multi-loss function which has been used in the literature to regulate deep networks \cite{xu2016multi}.

\subsection{Iterative Pruning for Projection Route Selection}
One important question in knowledge projection between the teacher and student networks is to determine which layers from the teacher network should be chosen as the knowledge layer and which layers from the students should be chosen for the injection layer. In this work, we propose to explore an iterative pruning and optimization scheme to select the projection route.

Assume that the teacher network $N_t$ and the student network $N_s$ have $L_t$ and $L_s$ layers, respectively. Candidate projection routes are depicted in Figure \ref{fig:route_selection}. We use only convolution layers as candidates for the knowledge and  injection layers. To satisfy the constraints on spatial size and receptive field, candidate knowledge projection routes are computed and denoted as $R_{i, j} \in \mathbb{G} $, where $i$ is the index of \textit{knowledge} layer in the teacher network, $j$ is the index of \textit{injection} layer in the student network, and $\mathbb{G}$ is the set of all candidate routes. We follow the procedure for computing the center of receptive field in \cite{lenc2015r} for calculating the size of receptive field in layer $L$:
\begin{equation}
\label{eq:receptive_field}
    S_L = \sum_{p=1}^{L}{(\prod_{q=1}^{p-1}{S_q}) (F_p - 1)},
\end{equation}
where $S_q$ and $F_p$ are the layer-wise stride and kernel size, assuming they are identical along $x$ and $y$ directions for simplicity. Routes with constrained receptive filed are kept after calculation with a small tolerance $\beta=0.2$:
\begin{equation}
\label{eq:tolerance}
    (1-\beta) \cdot S_i \le S_j \le (1+\beta) \cdot S_i.
\end{equation}
For example, in Figure \ref{fig:route_selection}, we have 
\begin{equation}
\{R_{1, 3}, R_{1, 4}, R_{2,5}\} \subset \mathbb{G}
\end{equation}
 and the rest routes in this figure are not valid due to mismatched spatial shapes. The idea of iterative pruning for the projection route selection is to traverse all possible routes with same training hyper-parameters, and determine the best route for \textit{knowledge-injection} pair on-the-fly. Specifically, we randomly initialize $|\mathbb{G}|$ KPNs according to each $R_{i, j}$. 
 
 Each KPN stores a student network $W_s$, knowledge projection parameter $W_{KP}$ and routing $R_{i, j}$, teacher network $W_t$ is shared across all KPNs to save computation and memory overhead. The target is to find the KPN setting with minimum joint loss
\begin{equation}
\label{eq:joint_loss}
\{W'_{s}, W'_{KP}, R'_{i, j}\} = \argmin_{\{W_{s}, W_{KP}, R_{i, j}\}}(\lambda \cdot \mathcal{L}_{KP} + \mathcal{L}_p).
\end{equation}
We assume that the pre-trained teacher network $W_t$ is responsible for guiding the training of a specifically designed student network $W_s$ which satisfies the computational complexity requirement. According to (\ref{eq:tolerance}), we can generate a list $\mathbb{L}$ of candidate KPNs. Each KPN is a copy of the designed student network $W_s$ with different projection routing $R_{i,j}$ and corresponding parameters $W_{KP}$. Within a period of $k$ epochs, the KPNs are optimized separately using Stochastic Gradient Descend to minimize the joint loss described in (\ref{eq:joint_loss}). Note that even though the optimization target is a joint loss, as depicted in Fig. \ref{fig:architecture}, the upper and bottom layers of the student network are receiving different learning targets from the teacher network and dataset distribution, respectively. At the end of $k$ epochs, the joint loss of each KPN computed on the validation dataset is used to determine which KPN to prune. The same procedure is applied on the remaining KPNs in the list $\mathbb{L}$ iteratively. This iterative pruning procedure is summarized in Algorithm \ref{alg:1}:

\begin{algorithm} \label{alg:1}
    \SetKwInOut{Input}{Input}
    \SetKwInOut{Output}{Output}

    %\underline{function Euclid} $(a,b)$\;
    \Input{List $\mathbb{L}$ of KPNs, as in form \{$W_{s, n}, W_{KP, n}, R_{i_n, j_n}$\}, where $n=1,...,|\mathbb{G}|$, and teacher network $W_{t}$}
    \Output{$W^*_s$, $W^*_{KP}$ and $R^*_{i, j}$}
    Configure all KPNs as initialization stage.\\
    \While{$|\mathbb{L}| > 1$}
      {
        \For{$k$ epochs}
        {
        	\For{Batch $x$ in Data}
        	{
     	        Forward teacher: $y_t \leftarrow \mu(x; W_k)$; \\
     	        \For{\{$W_{s}, W_{KP}, R_{i, j}$\} $ \in \mathbb{L}$}
     	        {	
     	        	Forward-backward w.r.t. $W_{s}, W_{KP}$;
     	        }
        	}
        	
        }
        \{$W'_{s}, W'_{KP}, R'_{i, j}$\} $\leftarrow \argmin(\lambda \cdot \mathcal{L}_{KP} + \mathcal{L}_p)$; \\
        Remove \{$W'_{s}, W'_{KP}, R'_{i, j}$\} in $\mathbb{L}$;
      }
      {
        return \{$W^*_{s}, W^*_{KP}, R^*_{i, j}$\} in $\mathbb{L}$;
      }
    \caption{Iterative pruning algorithm for projection route selection.}
\end{algorithm}

Only one KPN will \textit{survive} after the iterative pruning process. We continue the multi-stage training with or without adjusting the batch-size depending on the released memory size after sweeping out bad KPNs. The stopping criteria can either be plateau of validation accuracy or a pre-defined end epoch. 

%------------------------------------------------------------------------
\section{Experimental Results}
\label{sec:exp}

In this section, we provide comprehensive evaluations of our proposed method using three groups of benchmark datasets. Each group consists of two datasets, the large dataset $D_t$ used to train the teacher network and the smaller dataset $D_s$ used to train the student network. The motivation is that, in practical applications, we often need to learn a network to recognize or classify a relatively small number of different objects and the available training dataset is often small. We also wish the trained network to be fast and efficient. The large dataset is often available from existing research efforts, for example, the ImageNet. Both the large and the small datasets have the same image dimensions so that pre-trained models are compatible with each other in terms of shape. We use the existing teacher network model already trained by other researchers on the public dataset $D_t$. We compare various algorithms on the benchmark dataset $D_s$ where state-of-the-art results have been reported. Performance reports on small datasets are rare, thus we choose existing large famous benchmark datasets in following experiments, and aggressively reduce the size of training set to simulate the shortage of labeled data in real world scenarios.

\subsection{Network Training }
We have implemented our KPN framework using the MXNet \cite{chen2015mxnet}, a deep learning framework designed for both efficiency and flexibility. The dynamically generated computational graph in MXNet allows us to modify network structures during run time. The KPNs are trained on NVidia Titan X 12GB with CUDNN v5.1 enabled. Batch-sizes vary from 16 to 128 depending on the KPN group size. For all experiments, we train using the Stochastic Gradient Descend (SGD) with momentum 0.9 and weight decay 0.0001 except the knowledge projection layers. The weight decay for all knowledge projection layers is 0.001 in the initialization stage and 0 for the joint training stage. 40\% of iterations are used for the initialization stage, and the rest goes to be joint training stage. The weight controller parameter $\lambda$ for joint loss is set to be 0.6, and gradually decays to 0. The pruning frequency is 10000 and we also randomly revoke the initialization stage during joint training stage, to repetitively adjusting network guidance strength. 

For fine-tuning, we test with a wide variety of experimental settings. Starting from pre-trained networks, we adjust the last layer to fit to the new dataset, and randomly initialize the last layer. The reshaped network is trained with standard back-propagation with respect to labels on the new dataset, and \textit{unfreeze} one more layer from the bottom one at a time. The best result from all configurations was recorded. 
To make sure all networks are trained using the optimal hyper-parameter set, we extensively try a wide range of learning rates, and repeat experiments on the best parameter set for at least 5 times. The average performance of the best 3 runs out of 5 will be reported.  Data augmentation is limited to random horizontal flip if not otherwise specified.

%%%%%%%%%%%%%%%%%%%%%%%

\begin{table*}[ht]
\centering
\caption{CIFAR-10 accuracy and network capacity comparisons with state-of-the-art methods. Results using randomly sampled subsets from training data are also reported. Number of network parameters are calculated based on reports in related work.}
\label{tab:cifar}
\begin{tabular}{l|l|l|l|l|l|l}
\specialrule{.2em}{.1em}{.1em} 
\multirow{2}{*}{Methods} & \multicolumn{4}{l|}{Accuracy with Different $S_{T}$}     & \multirow{2}{*}{$N_{Para}$} & \multirow{2}{*}{$N_{MA}$} \\ \cline{2-5}
                         & 50000    & 5000   & 1000  & 500   &     &                       \\ 
 \specialrule{.15em}{.05em}{.05em}
Maxout \cite{goodfellow2013maxout}         & 90.18  & - & -  & -       & 9M       & 379M                  \\ \hline
FitNets-11 \cite{romero2014fitnets}            & 91.06  & - & -  & -         & 0.86M        & 53M              \\ \hline
FitNets \cite{romero2014fitnets}           & 91.61  & - & -  & -          & 2.5M       & 107M                \\ \hline
GP CNN \cite{lee2016generalizing}            & 93.95 & - & -  & -        & 3.5M           & 362M               \\ \hline
ALL-CNN-C \cite{springenberg2014striving}        & 92.7 & - & -  & -  & 1.0M        & 257M                 \\ \hline
Good Init \cite{mishkin2015all}         & 94.16  & - & -  & -        & 2.5M         & 166M                 \\
\specialrule{.15em}{.05em}{.05em}     
ResNet-50 slim          & 87.53    & 71.92  & 55.86 & 48.17 & \textbf{0.27M} & \textbf{31M}                  \\ \hline
ResNet-38             & 90.86    & 75.28  & 61.74 & 51.62 & 3.1M   & 113M                    \\ \hline
ResNet-38 fine-tune          & 91.15  & 89.61  & 86.26 & 83.45 & 3.1M  & 113M               \\ \hline
Our method                 & \textbf{92.37}    & \textbf{90.35}  & \textbf{88.73} & \textbf{87.61} & \textbf{0.27M} & \textbf{31M}                  \\ \specialrule{.1em}{.05em}{.05em}  
\end{tabular}
\end{table*}

\subsection{Results on the CIFAR-10 Dataset}

We first evaluate the performance of our method on the CIFAR-10 dataset guided by a teacher network pre-trained on  CIFAR-100 dataset. The CIFAR-10 and CIFAR-100 datasets \cite{krizhevsky2009learning} have 60000 32$\times$32 color images with 10 and 100 classes, respectively. They were both split into 50K-10K sets for training and testing. To validate our approach, we trained a 38-layer Resnet on the CIFAR-100 as reported in \cite{he2016deep}, and use it to guide a 50-layer but significantly \textit{slimmer} Resnet on the CIFAR-10. We augment the data using random horizontal flip and color jittering. Table \ref{tab:cifar} summarizes the results, with comparisons against the state-of-the-art results which cover a variety of optimization techniques including Layer-sequential unit-variance initialization \cite{mishkin2015all}, pooling-less \cite{springenberg2014striving}, generalized pooling \cite{lee2016generalizing} and maxout activation \cite{goodfellow2013maxout}.  We choose different sizes $S_T$ of the training set and list the accuracy. For network complexity, we compute its number of model parameters $N_{Para}$ and the number of multiplication and additions $N_{MA}$ needed for the network inference.
It should be noted that for methods in the literature we do not have  their accuracy results on down-sized training sets.

We do not apply specific optimization techniques used in the state-of-the-art methods due to some structures not reproducible in certain conditions. To compare, we trained a standard 38-layer Residue Network, a 50-layer slimmer version of ResNet (each convolutional layer is half the capacity of the vanilla ResNet) and a fine-tuned model of 38-layer ResNet (from CIFAR-100) on CIFAR-10 with different amount of training samples.  With all 50000 training data, our proposed method outperforms direct training and best fine-tuning results and still match the state-of-the-art performance. 
We believe the performance gain specified in \cite{lee2016generalizing, mishkin2015all} can be also applied to our method, \ie, ensemble of multiple techniques could achieve better performance. The proposed KPN method has improved the accuracy by up to 1.2\% while significantly reducing the network size by about 11 times, from 3.1M network parameters to 273K parameters. It also demonstrated strong robustness against aggressive reduction of labeled training samples.

\begin{figure}[h]
\begin{center}
%\fbox{\rule{0pt}{2in} \rule{0.7\linewidth}{0pt}}
\includegraphics[width=0.95\linewidth]{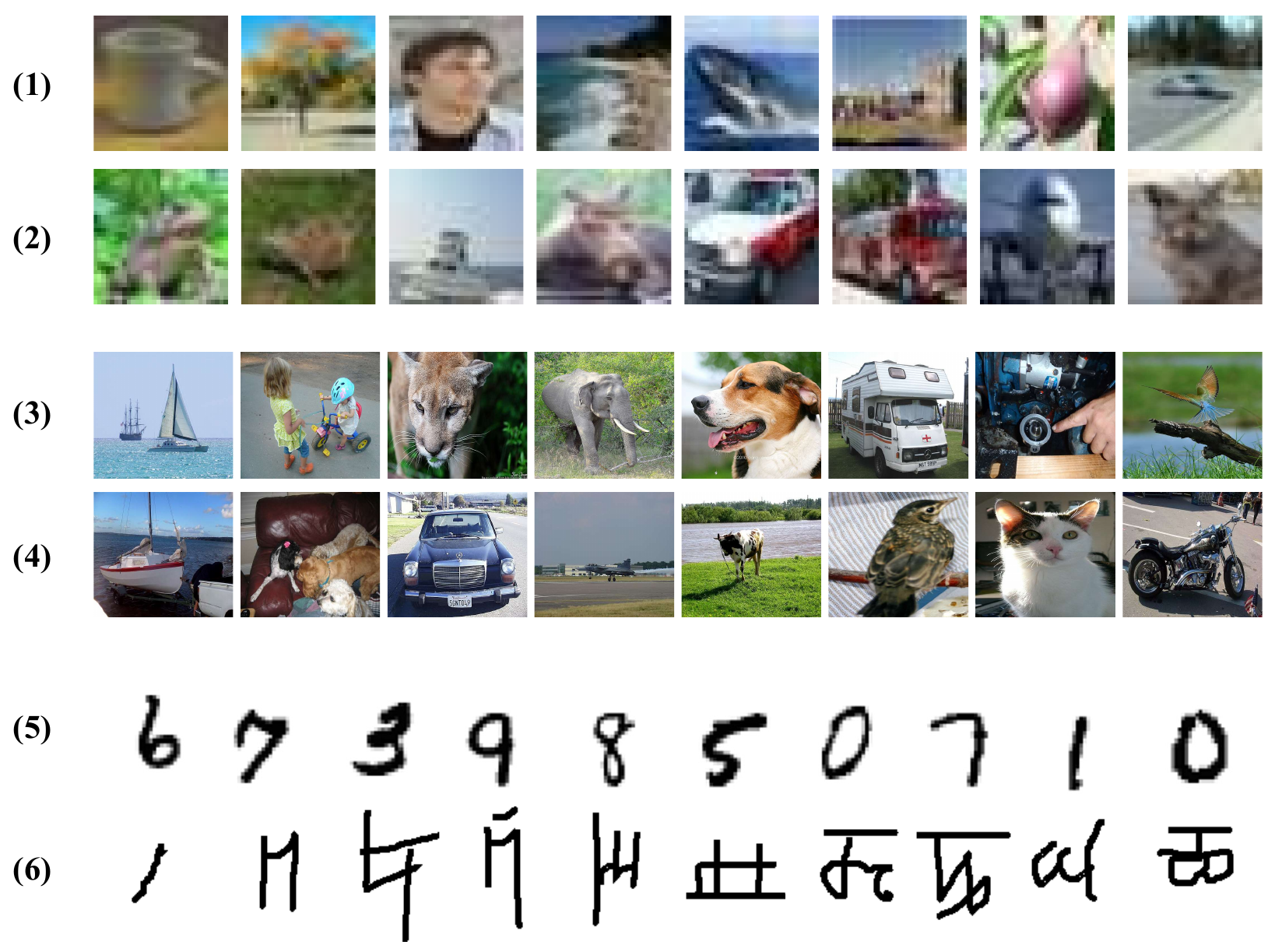}
\end{center}
\caption{(1)(2): CIFAR-100/10 sample images; (3): Imagenet 2012; (4) Pascal VOC 2007; (5) MNIST; (6) Omniglot;}
\label{fig:samples}
\end{figure}

\subsection{Results on the Pascal VOC 07 Dataset}

\begin{table*}[ht]
\centering

\caption{PASCAL VOC 2007 test object classification performances comparison. Results using randomly sampled subsets from training data are also reported. Number of convolution layer parameters are listed for fair comparison based on reports in related work.}
\begin{tabular}{l|l|l|l|l|l}
\specialrule{.2em}{.1em}{.1em} 
\multirow{2}{*}{Methods}     & \multicolumn{3}{l|}{Accuracy at Different $S_T$}       & \multirow{2}{*}{$N_{Para}$} & \multirow{2}{*}{$N_{MA}$} \\ \cline{2-4}
                             & 5011 & 1000 & 200         &             &                          \\
                             \specialrule{.1em}{.05em}{.05em}
Chatfield \etal \cite{chatfield2014return}               & 82.4 & - & -           & 6.5M       & 2483M           \\ \hline

VGG16+SVM \cite{simonyan2014very}                    & 89.3 & - & -            & 14.7M          & 15470M             \\ \hline
VGG19+SVM \cite{simonyan2014very}                    & 89.3 & - & -             & 21.8M      & 15470M                  \\ \hline
HCP-VGG \cite{wei2016hcp}                     & 90.9 & - & -             & 14.7M      & 15470M                  \\ \hline
FisherNet-VGG16 \cite{tang2016deep}             & \textbf{91.7} & - & -             & 14.7M       & 15470M                  \\ \specialrule{.15em}{.05em}{.05em}
VGG16 standard BP                    & 83.5           & 65.2 & \textless30 & 14.7M       & 15470M                 \\ \hline
Fine-tune VGG16 last layer (softmax) & 89.6           & 87.4 & 85.7        & 14.7M        & 15470M                 \\ \hline
Fine-tune VGG16 2+ learnable layers           & 90.2           & 86.3 & 82.8        & 14.7M        & 15470M                \\ \hline
Our method                     & \textbf{91.2}           & \textbf{88.4} & \textbf{86.5}        & \textbf{8M}     & \textbf{3361M}                 \\ \specialrule{.1em}{.05em}{.05em}  
 \end{tabular}
\label{tab:pascal}
\end{table*}

We evaluate the proposed method on the PASCAL Visual Object Classes Challenge(VOC) dataset \cite{everingham2010pascal} with a VGG-16 model \cite{simonyan2014very} pre-trained on the ILSVRC 2012 dataset \cite{ILSVRC15}. The pre-training usually takes several weeks, thus we downloaded and converted the teacher network from the Caffe model available online. We compare our method with state-of-the-art results obtained on this dataset in the literature, 
including the VGG16+SVM method \cite{simonyan2014very},
the segment hypotheses based multi-label HCP-VGG method \cite{wei2016hcp}, and the 
FisherNet-VGG16 method \cite{tang2016deep} which encodes CNN feature with fisher vector. 
These papers have reported results on the original whole dataset with 5011 images.
To test the learning capability of the network on smaller datasets with reduced samples, we also implement the fine-tuning method. We try different combination of network update scheme and learning parameters and use the best result for performance comparison with our method. 
We conducted our experiments on the entire training set with 5011 images and \textit{test} set with 4952 images. In addition, we randomly sample 50 and 10 images from each class, generating two small datasets with 1000 and 200 training images, respectively. 
The results are summarized in Table \ref{tab:pascal}.   We list the test accuracy of the network for each configuration.
We compute the corresponding complexity of the network, including the number of model parameters $N_{Para}$ and the number of multiplication and additions $N_{MA}$.
It should be noted that for methods in the literature we do not have  their accuracy results on down-sized training sets.
It can be seen that our proposed method outperforms standard training and fine-tuning by a large margin while reducing the model size by 2 times and improving the inference speed by 4.6 times.

\subsection{Results on the Ommniglot Dataset}
We are interested in how the proposed KPN method works on very small datasets, for example, the Ommniglot handwritten recognition dataset. 
The MNIST \cite{lecun1998gradient} is a famous handwritten digits dataset, consists of 60000 training images and 10000 test images, 28x28x1 in size, organized into 10 classes. The Omniglot \cite{lake2015human} is a similar but much smaller dataset, containing 1623 different handwritten characters from 50 alphabets. Each of the 1623 characters was drawn online via Amazon's Mechanical Turk by 20 different people. All images are binarized and resized to 28$\times$28$\times$1 with no further data augmentation. We use all 70000 images from MNIST for training a 5-layer Maxout convolutional model as the teacher network $N_t$ as proposed in \cite{goodfellow2013maxout}. We report experimental results of various algorithms across a wide range of number of training examples, from 19280 to merely 1000, shown in Table \ref{tab:omniglot}. Note that we use class dependent shuffling to randomly select training subsets, which is critical to avoid unbalanced class distribution in Omniglot due to the limited number of samples for each class. 
We can see that the proposed KPN is able to reduce the error rate by 1.1-1.3\%. 
Table \ref{tab:omniglot} also provides some interesting insights of how models are transferred to different tasks. 
First, the fine-tuning methods are all affected by the number of learnable parameters and training samples. Smaller training set will result in significant over-fitting, thus breaking the fragile co-adaptation between layers. If the training set is large enough, the number of learnable parameters are positively related to the performance. This phenomenon is also discussed in \cite{yosinski2014transferable}, where transferring knowledge from the pre-trained model to an exactly same network is extensively tested. 

% \usepackage{multirow}
\begin{table}[h]
\centering
\caption{Test error rate comparisons between experimental settings and baseline methods. }
\label{tab:omniglot}
\begin{tabular}{l|l|l|l}
\specialrule{.2em}{.1em}{.1em} 
\multirow{2}{*}{Methods} & \multicolumn{3}{l}{Error Rates at Different $S_T$} \\ \cline{2-4} 
                         & 19280      & 5000      & 1000      \\ \hline
                     \specialrule{.15em}{.05em}{.05em}  
Deep CNN  \cite{lake2015human}               & 13.5\% & - & -          \\ \hline
Deep Siamese CNN  \cite{lake2015human}        & 8.0 \%& - & -             \\ \specialrule{.15em}{.1em}{.1em} 
Large CNN standard BP                 & 9.3\%        & 12.9\%      & 19.4\%      \\ \hline
Small CNN standard BP                 & 12.1\%       & 18.5\%      & 23.8\%      \\ \hline
Fine-tuned from MNIST              & 6.8\%        & 7.4\%       & 9.2\%       \\ \hline
Our method       & \textbf{5.9\%}        & \textbf{6.6\%}       & \textbf{7.9\%}       \\ \specialrule{.2em}{.1em}{.1em}    
\end{tabular}
\end{table}

\begin{figure}[h]
\begin{center}
   \includegraphics[width=0.95\linewidth]{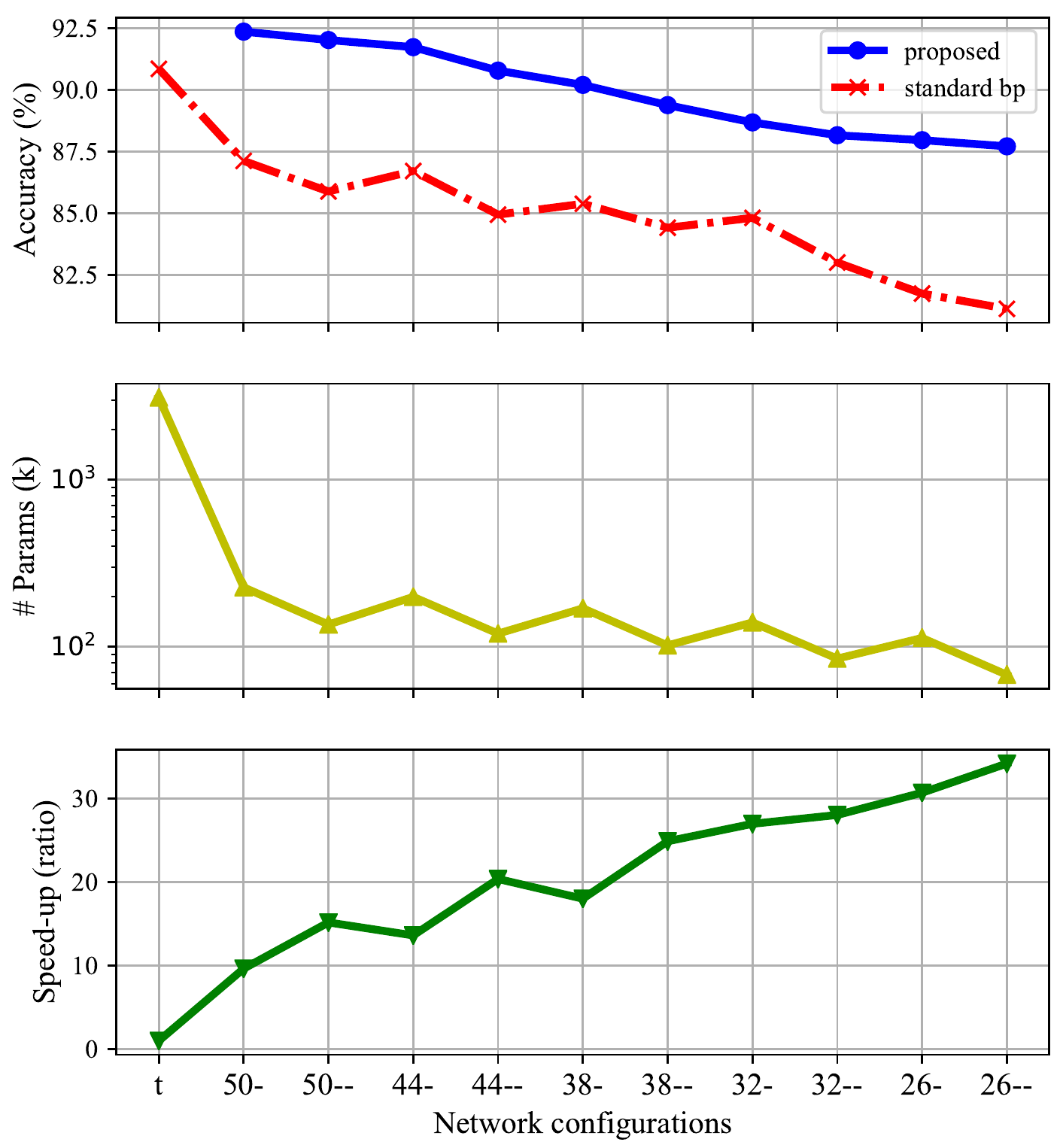}
\end{center}
   \caption{Network capacity and performance analysis. Top: test accuracies with proposed KPN and normal training with standard back-propagation; Middle: number of parameters ($\times 10^3$), note that the y-axis is in logarithmic scale; Bottom: actual inference speed up ratio with respect to Resnet-50. Network notations: $t$ is teacher network, $N$- denotes slim network with $N$ layers, similarly, $N$ layer slimmer network is denoted as $N$-\ -.}
\label{fig:tradeoff}
\end{figure}

\begin{table*}[]
\centering
\caption{Network configurations for extensive benchmarks on Omniglot dataset. $N$- denotes slim network with $N$ layers, similarly, $N$ layer slimmer network is denoted as $N$-\ -. Note that $1 \times 1$ adaptive convolutions for residue modules are not included in this table.}
\label{tab:net_config}
\resizebox{0.9\linewidth}{!}{%
\begin{tabular}{l|l|l|l|l|l|l|l|l|l|l|l}
\specialrule{.2em}{.1em}{.1em} 
\# Layers      & 50     & 50-   & 50-\ -  & 44-   & 44-\ -  & 38- & 38-\ - & 32- & 32-\ - & 26- & 26-\ - \\ \hline
\specialrule{.1em}{.05em}{.05em}  
Conv$3\times 3$ /s1    & 16     & 16    & 16    & 16    & 16    & 16  & 16   & 16  & 16   & 16  & 16   \\ \hline
ResConv$3\times 3$ /s2 & 32 $\times 16$  & 32 $\times 16$ & 16 $\times 16$ & 32 $\times 14$ & 16 $\times 14$ & 32 $\times 12$   & 16 $\times 12$ & 32 $\times 10$   & 16 $\times 10$  & 32 $\times 8$  &  16 $\times 8$    \\ \hline
ResConv$3\times 3$ /s1 & 64 $\times 16$  & 32 $\times 16$ & 32 $\times 16$ & 32 $\times 14$ & 32 $\times 14$ & 32 $\times 12$   & 32 $\times 12$ & 32 $\times 10$   & 32 $\times 10$  & 32 $\times 8$  & 32 $\times 8$   \\ \hline
ResConv$3\times 3$ /s2 & 128 $\times 16$ & 64 $\times 16$ & 48 $\times 16$ & 64 $\times 14$ & 48 $\times 14$ & 64 $\times 12$   & 48 $\times 12$ & 64 $\times 10$  &  48 $\times 10$ &  64 $\times 8$ & 48 $\times 8$   \\ \hline
Conv$3\times 3$ /s1     & 256    & 128   & 96    & 128   & 96    & 128   & 96   & 128   & 96   & 128   & 96   \\
\specialrule{.2em}{.1em}{.1em} 
\end{tabular}
}
\end{table*}

\begin{figure}[h]
\begin{center}
   \includegraphics[width=0.95\linewidth]{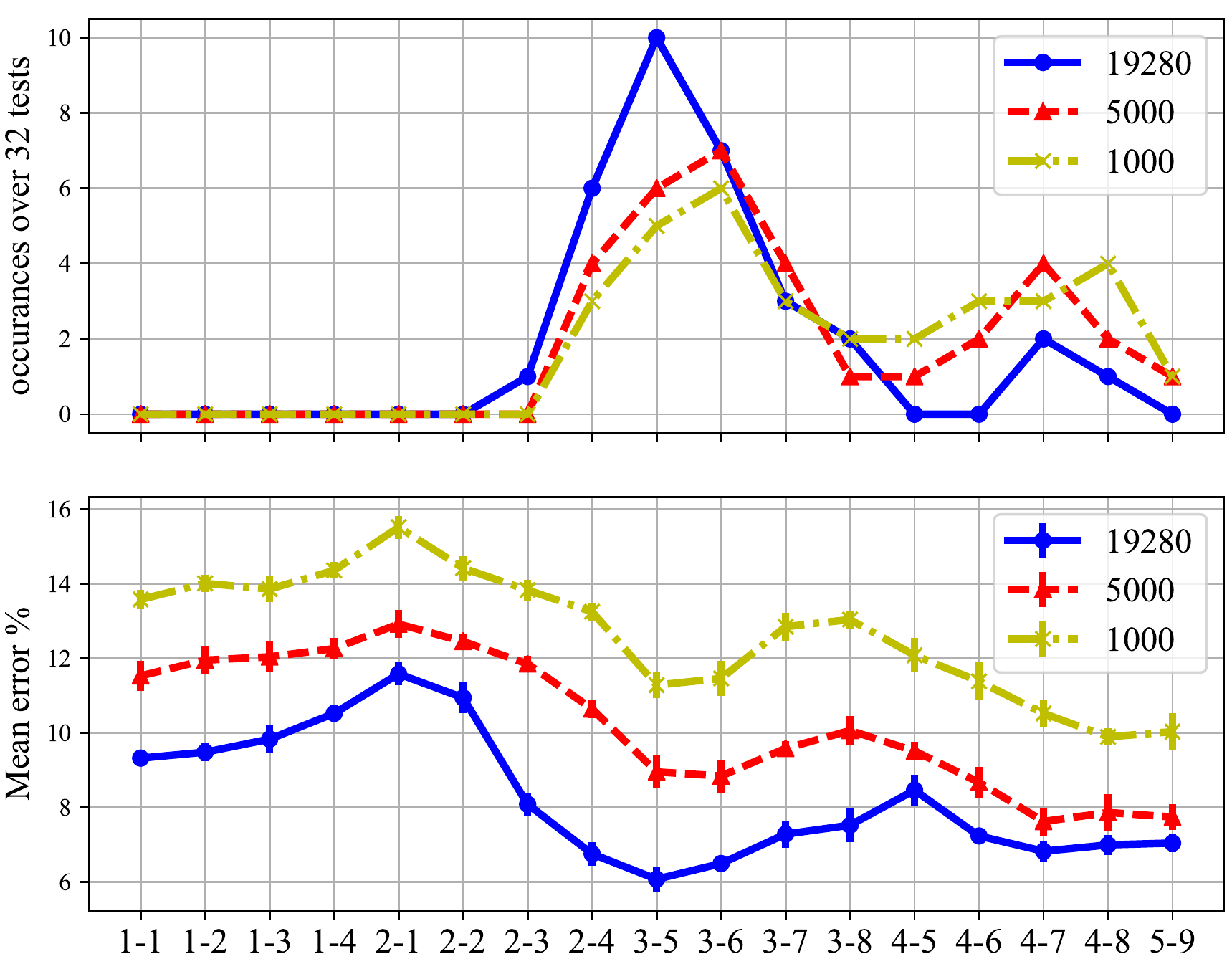}
\end{center}
   \caption{Iterative pruning analysis. Top: occurrences of projection route $t$-$s$ over 32 standalone tests. Bottom: mean classification error of projection route $t$-$s$ by disable iterative pruning. $t$-$s$: network with knowledge layer $t$ from teacher to injection layer $s$ from student.}
\label{fig:pruning}
\end{figure}

\subsection{Algorithm Parameter Analysis}
In this section, we study how the performance of the our method is impacted by the selection of major parameters. 

(1) \textbf{Trade-off between Performance and Efficiency}. To evaluate how the size of network affects the performance, we measure the test accuracy, number of parameters, and network speed up ratio of various student networks on the CIFAR-10 dataset. Figure \ref{fig:tradeoff} shows the results. Student networks are designed based on  a multi-layer Resnet denoted as $N$- or $N$- -, where $N$ is the number of layers, - and - - indicate it's a slim or slimmer version of Resnet. The detailed network configurations are listed in Table \ref{tab:net_config}. As expected, deeper and slimmer networks are more difficult to train with limited training data. However, with proposed method enabled, the depth is beneficial, and networks are less suffered from performance drop. Impressively, we could obtain a model which is 34 times faster using less than 2\% parameters, with about 3\% accuracy loss, compared to the teacher network.

(2) \textbf{Analysis of Iterative Pruning for Automatic Route Selection}. The knowledge projection route is critical for the network training and test performance. Intuitively, the projection route should not be too shallow or too deep. Shallow layers may contain only low-level texture features, while deep layers close to output may be too task specific. To study how the iterative pruning works during training, we record the pruning results and compare them with respect to manually defined projection routes, shown in Figure \ref{fig:pruning}. We can see that the statistics of survival projection routes is highly correlated to the training accuracy, which is evaluated by manually defining projection route from $t$ to $s$ and disabling iterative pruning during training. The result also indicates that choosing the middle layers for projection is potentially better. Reducing the size of training data also affects the pruning results. This might relate to the difficulty of fitting knowledge projection layer to the target domain when very limited data is presented. As a result, projection layers  tend to appear more on very deep layers close to the output, so that the penalty from adaptation loss will not dominate. The bottom line is, even though the iterative pruning method is a random optimization process, it is reliably producing satisfactory results.

\subsection{Discussion and Future Work}
Our KPN is designed in a highly modular manner. The training of projection layers is removed during actual network testing, and the network capacity is highly configurable for performance/speed trade-off. This KPN method can be easily extended to other problems such as object detection, object segmentation, and pose estimation by replacing softmax loss layer used in the  classification problems. Since the deployed network is a pure standard network, another research direction is to apply KPN as a building block in traditional model compression techniques to reshape the network in a new perspective. Although we have focused on the advantage of KPN with thinner networks on smaller datasets, there are  potential benefits to apply KPN on large network and relatively large datasets, for example, performance oriented situations where speed is not an issue.

\section{Conclusion}
\label{sec:con}
We have developed a novel knowledge projection framework for deep neural networks the address the issues of domain adaptation and model compression in training simultaneously. We exploit the distinctive general features produced by the  teacher network trained on large dataset, and use a learned matrix to project them into domain relevant representations to be used by the student network. A smaller and faster student network is trained to minimize joint loss designed for domain adaptation and knowledge distillation simultaneously. Extensive experimental results have demonstrated that our unified training framework provides an effective way to obtain fast high-performance neural networks on small datasets
with limited labeled samples.

% or


%\appendices

%\section{}

%\section*{Acknowledgment}


\ifCLASSOPTIONcaptionsoff
  \newpage
\fi



\bibliographystyle{IEEEtran}
\bibliography{background}

\end{document}